\documentclass[10pt,twocolumn,letterpaper]{article}

\usepackage{cvpr}
\usepackage{times}
\usepackage{epsfig}
\usepackage{graphicx}
\usepackage{amsmath}
\usepackage{amssymb}
\usepackage{breqn}
\usepackage{float}
\usepackage{algorithm}
\usepackage{algorithmic}
\usepackage{amsmath}
\usepackage{capt-of}
\usepackage{pdfpages}
\usepackage{enumerate}
\usepackage{footnote}
\usepackage{multirow}
\usepackage{makecell}

\usepackage[pagebackref=true,breaklinks=true,letterpaper=true,colorlinks,bookmarks=false]{hyperref}

\cvprfinalcopy

\def\cvprPaperID{3662} 

\begin{document}

\title{StarGAN: Unified Generative Adversarial Networks \\ for Multi-Domain Image-to-Image Translation}

\author{Yunjey Choi\textsuperscript{1,2} \hspace{.04in} Minje Choi\textsuperscript{1,2} \hspace{.04in} Munyoung Kim\textsuperscript{2,3} \hspace{.04in} Jung-Woo Ha\textsuperscript{2} 
\vspace{.05in}
\hspace{.04in} Sunghun Kim\textsuperscript{2,4} \hspace{.04in} Jaegul Choo\textsuperscript{1,2}\\
\textsuperscript{1}\thinspace Korea University \hspace{.1in} \vspace{.03in}\textsuperscript{2}\thinspace Clova AI Research, NAVER Corp.\\
\textsuperscript{3}\thinspace The College of New Jersey \hspace{.1in} \textsuperscript{4}\thinspace Hong Kong University of Science \& Technology
\vspace{-.1in}
}

\twocolumn[{
\renewcommand\twocolumn[1][]{#1}
\maketitle
\begin{center}
    \centering
    \includegraphics[width=1.0\textwidth]{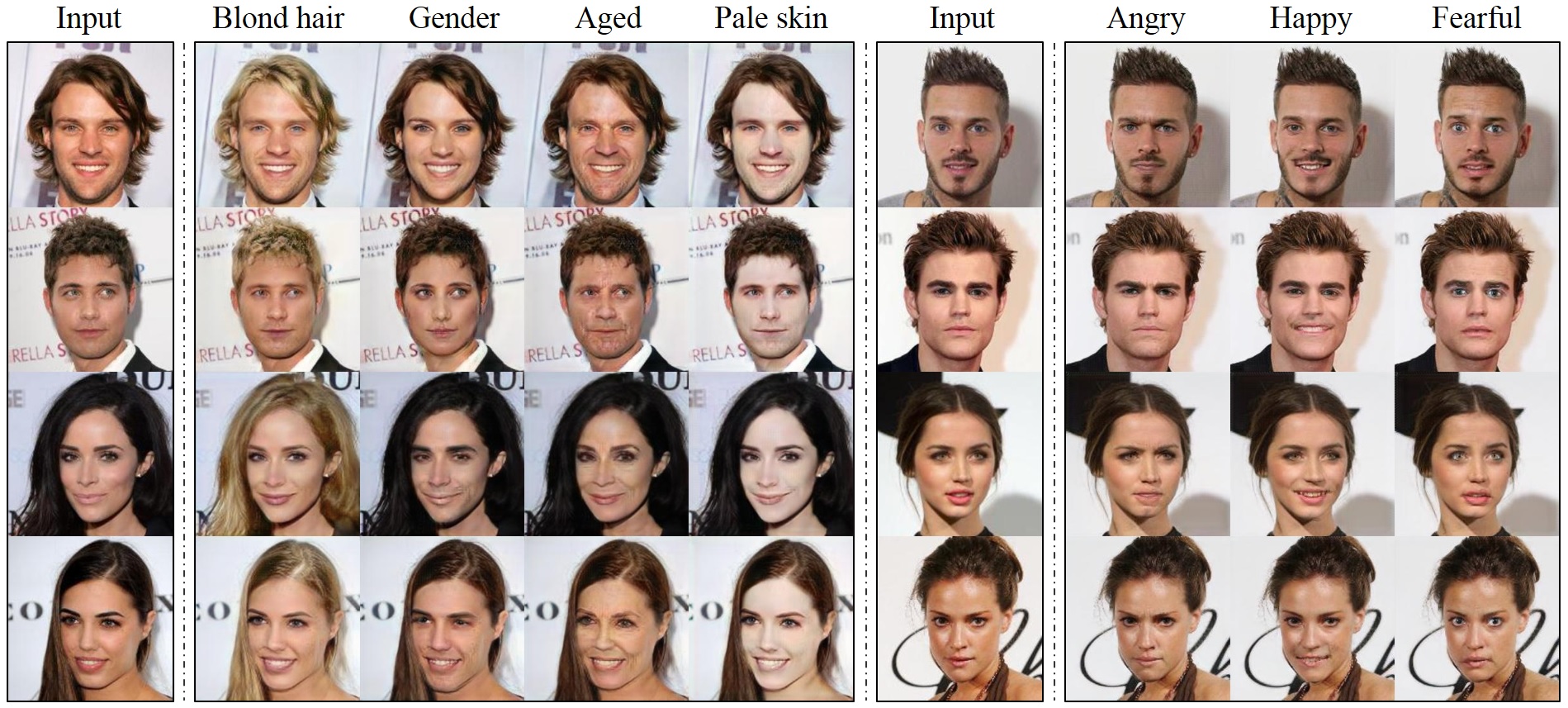}
    \captionof{figure}{Multi-domain image-to-image translation results on the CelebA dataset via transferring knowledge learned from the RaFD dataset. The first and sixth columns show input images while the remaining columns are images generated by StarGAN. Note that the images are generated by a single generator network, and facial expression labels such as angry, happy, and fearful are from RaFD, not CelebA.}
    \label{celeb}
\end{center}
}]

\begin{abstract}
\vspace{-.2in}
Recent studies have shown remarkable success in image-to-image translation for two domains. However, existing approaches have limited scalability and robustness in handling more than two domains, since different models should be built independently for every pair of image domains. To address this limitation, we propose StarGAN, a novel and scalable approach that can perform image-to-image translations for multiple domains using only a single model. Such a unified model architecture of StarGAN allows simultaneous training of multiple datasets with different domains within a single
network. This leads to StarGAN's superior quality of translated images
compared to existing models as well as the novel capability of flexibly translating an input image to any desired target domain. We empirically demonstrate the effectiveness of our approach on a facial attribute transfer and a facial expression synthesis tasks.
\end{abstract}

\vspace{-.2in}
\section{Introduction}

The task of image-to-image translation is to change a particular aspect of a given image to another, e.g., changing the facial expression of a person from smiling to frowning (see Fig.\thinspace\ref{celeb}). This task has experienced significant improvements following the introduction of generative adversarial networks (GANs), with results ranging from changing hair color~\cite{kim2017learning}, reconstructing photos from edge maps~\cite{Isola_2017_CVPR}, and changing the seasons of scenery images~\cite{zhu2017unpaired}. 

Given training data from two different domains, these models learn to translate images from one domain to the other. We denote the terms \textit{attribute} as a meaningful feature inherent in an image such as hair color, gender or age, and \textit{attribute value} as a particular value of an attribute, e.g., black/blond/brown for hair color or male/female for gender. We further denote \textit{domain} as a set of images sharing the same attribute value. For example, images of women can represent one domain while those of men represent another. 

Several image datasets come with a number of labeled attributes. For instance, the CelebA\cite{liu2015faceattributes} dataset contains 40 labels related to facial attributes such as hair color, gender, and age, and the RaFD \cite{langner2010presentation} dataset has 8 labels for facial expressions such as `happy', `angry' and `sad'. These settings enable us to perform more interesting tasks, namely \textit{multi-domain image-to-image translation}, where we change images according to attributes from multiple domains. The first five columns in Fig.\thinspace\ref{celeb} show how a CelebA image can be translated according to any of the four domains, `blond hair', `gender', `aged', and `pale skin'. We can further extend to training multiple domains from different datasets, such as jointly training CelebA and RaFD images to change a CelebA image's facial expression using features learned by training on RaFD, as in the rightmost columns of Fig.\thinspace\ref{celeb}. 

\begin{figure}[t]
\centering
\centerline{\includegraphics[width=1.0\linewidth]{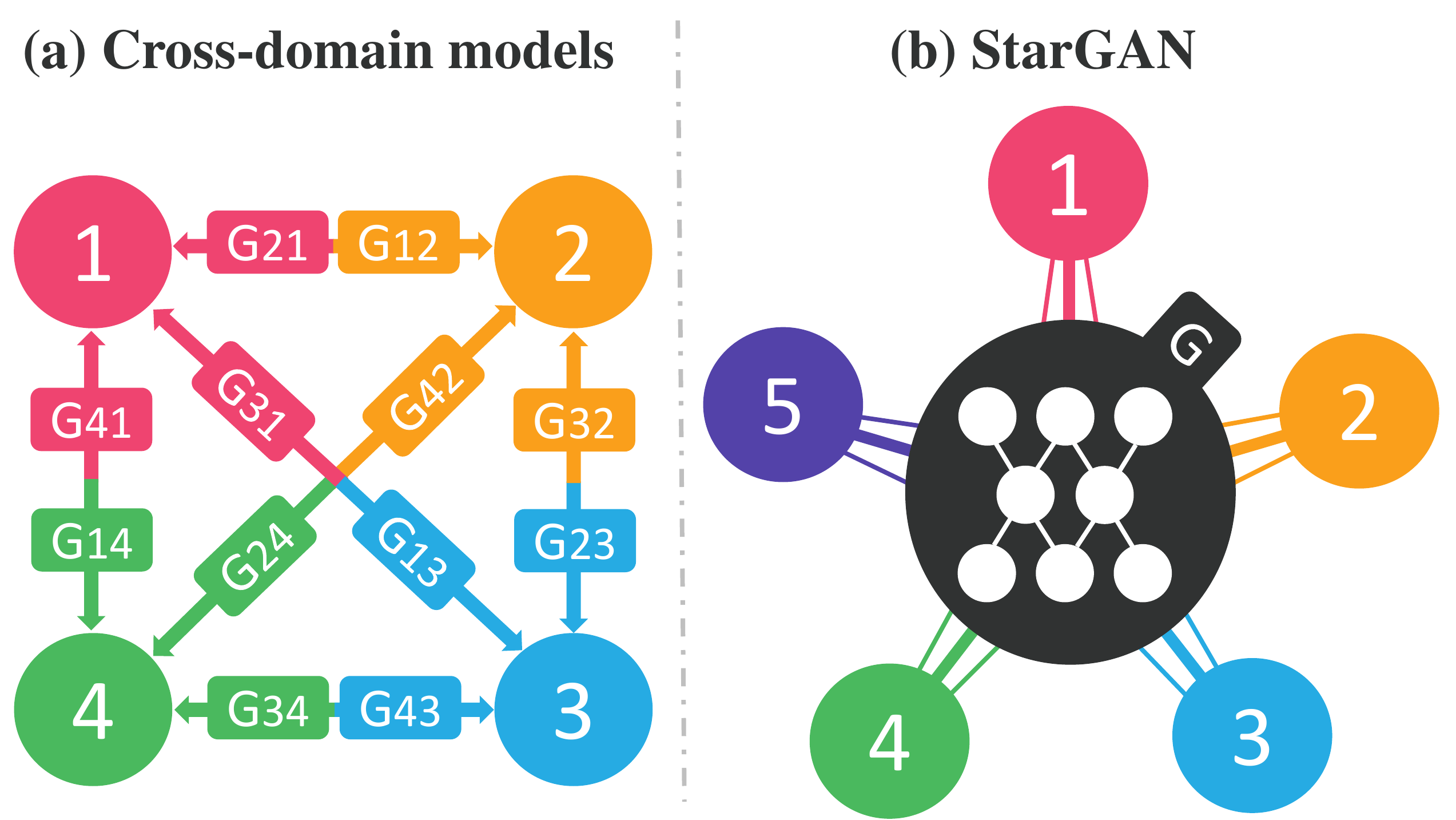}}
\caption{Comparison between cross-domain models and our proposed model, StarGAN. (a) To handle multiple domains, cross-domain models should be built for every pair of image domains. (b) StarGAN is capable of learning mappings among multiple domains using a single generator. The figure represents a star topology connecting multi-domains.}
\label{motiv}
\end{figure}

However, existing models are both inefficient and ineffective in such multi-domain image translation tasks. Their inefficiency results from the fact that in order to learn all mappings among $k$ domains, $k(k\mathbb{-}1)$ generators have to be trained. Fig.\thinspace\ref{motiv}\thinspace(a) illustrates how twelve distinct generator networks have to be trained to translate images among four different domains. Meanwhile, they are ineffective that even though there exist global features that can be learned from images of all domains such as face shapes, each generator cannot fully utilize the entire training data and only can learn from two domains out of $k$. Failure to fully utilize training data is likely to limit the quality of generated images. Furthermore, they are incapable of jointly training domains from different datasets because each dataset is \textit{partially labeled}, which we further discuss in Section \ref{section 3.2}.


As a solution to such problems we propose StarGAN, a novel and scalable approach capable of learning mappings among multiple domains. As demonstrated in Fig.\thinspace\ref{motiv}\thinspace(b), our model takes in training data of multiple domains, and learns the mappings between all available domains using only a single generator. The idea is simple. Instead of learning a fixed translation (e.g., black-to-blond hair), our generator takes in as inputs both image and domain information, and learns to flexibly translate the image into the corresponding domain. We use a label (e.g., binary or one-hot vector) to represent domain information. During training, we randomly generate a target domain label and train the model to flexibly translate an input image into the target domain. By doing so, we can control the domain label and translate the image into any desired domain at testing phase. 

We also introduce a simple but effective approach that enables joint training between domains of different datasets by adding a mask vector to the domain label. Our proposed method ensures that the model can \textit{ignore} unknown labels and \textit{focus} on the label provided by a particular dataset. In this manner, our model can perform well on tasks such as synthesizing facial expressions of CelebA images using features learned from RaFD, as shown in the rightmost columns of Fig.\thinspace\ref{celeb}. As far as our knowledge goes, our work is the first to successfully perform multi-domain image translation across different datasets.

\vspace{4px}

Overall, our contributions are as follows: 
\begin{enumerate}
\item[$\bullet$] We propose StarGAN, a novel generative adversarial network that learns the mappings among multiple domains using only a single generator and a discriminator, training effectively from images of all domains.
\item[$\bullet$] We demonstrate how we can successfully learn multi-domain image translation between multiple datasets by utilizing a mask vector method that enables StarGAN to control all available domain labels.
\item[$\bullet$] We provide both qualitative and quantitative results on facial attribute transfer and facial expression synthesis tasks using StarGAN, showing its superiority over baseline models.

\end{enumerate}


\begin{figure*}[t]
\centering
\centerline{\includegraphics[width=1.0\linewidth]{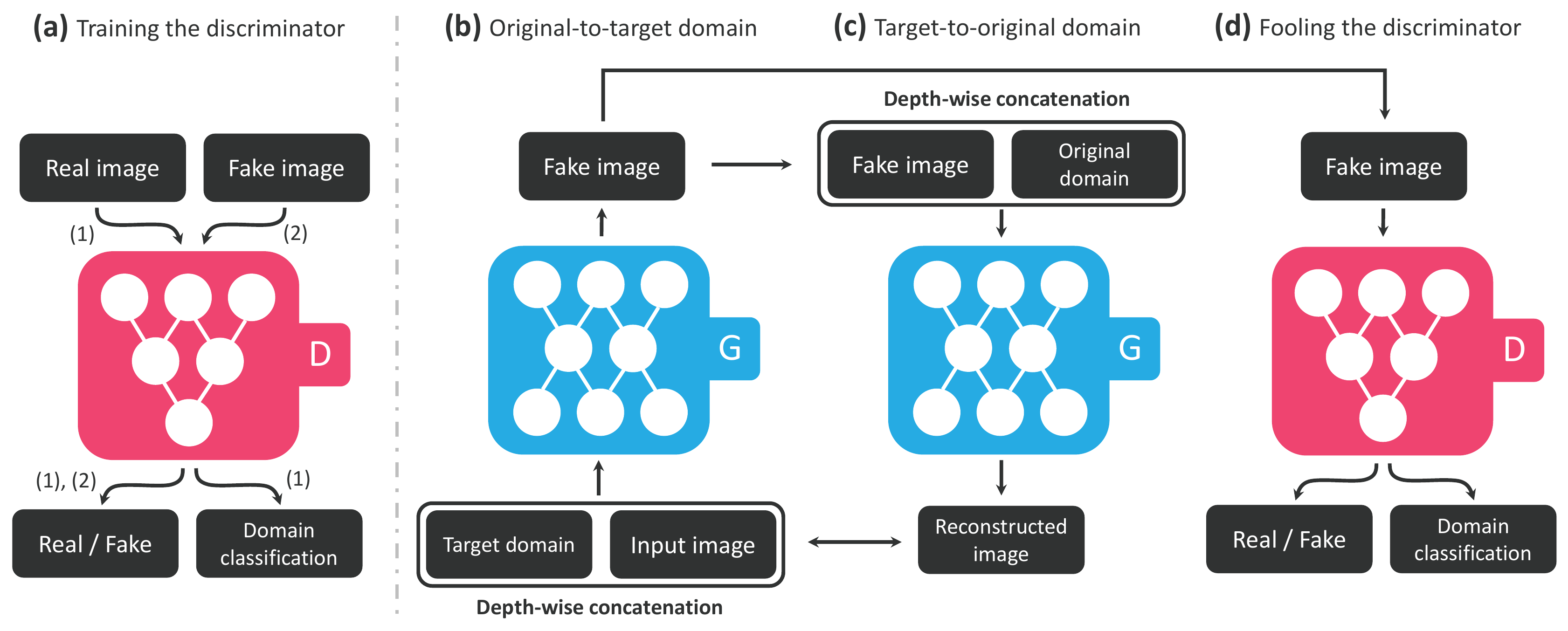}}
\caption{Overview of StarGAN, consisting of two modules, a discriminator $D$ and a generator $G$. \textbf{(a)} $D$ learns to distinguish between real and fake images and classify the real images to its corresponding domain. \textbf{(b)} $G$ takes in as input both the image and target domain label and generates an fake image. The target domain label is spatially replicated and concatenated with the input image. \textbf{(c)} $G$ tries to reconstruct the original image from the fake image given the original domain label. \textbf{(d)} $G$ tries to generate images indistinguishable from real images and classifiable as target domain by $D$.}
\label{model}
\end{figure*}

\vspace{-5px}
\section{Related Work} \label{related work}

\noindent\textbf{Generative Adversarial Networks.} Generative adversarial networks (GANs) \cite{goodfellow2014generative} have shown remarkable results in various computer vision tasks such as image generation \cite{Huang_2017_CVPR,radford2015unsupervised,zhao2016energy,karras2017progressive}, image translation \cite{Isola_2017_CVPR,kim2017learning,zhu2017unpaired}, super-resolution imaging \cite{Ledig_2017_CVPR}, and face image synthesis \cite{kim2017unsupervised,li2016deep,Shen_2017_CVPR,Zhang_2017_CVPR}. A typical GAN model consists of two modules: a discriminator and a generator. The discriminator learns to distinguish between real and fake samples, while the generator learns to generate fake samples that are indistinguishable from real samples. Our approach also leverages the adversarial loss to make the generated images as realistic as possible.

\medskip

\noindent\textbf{Conditional GANs.} GAN-based conditional image generation has also been actively studied. Prior studies have provided both the discriminator and generator with class information in order to generate samples conditioned on the class ~\cite{mirza2014conditional,odena2016semi,odena2016conditional}. Other recent approaches focused on generating particular images highly relevant to a given text description ~\cite{reed2016generative,zhang2016stackgan}. The idea of conditional image generation has also been successfully applied to domain transfer \cite{kim2017learning,taigman2016unsupervised}, super-resolution imaging\cite{Ledig_2017_CVPR}, and photo editing \cite{brock2016neural,Shu_2017_CVPR}.
In this paper, we propose a scalable GAN framework that can flexibly steer the image translation to various target domains, by providing conditional domain information. 

\medskip

\noindent \textbf{Image-to-Image Translation.} Recent work have achieved impressive results in image-to-image translation \cite{Isola_2017_CVPR, kim2017learning,liu2017unsupervised,zhu2017unpaired}. For instance, pix2pix \cite{Isola_2017_CVPR} learns this task in a supervised manner using cGANs\cite{mirza2014conditional}. It combines an adversarial loss with a L1 loss, thus requires paired data samples. To alleviate the problem of obtaining data pairs, unpaired image-to-image translation frameworks \cite{kim2017learning,liu2017unsupervised,zhu2017unpaired} have been proposed.
UNIT \cite{liu2017unsupervised} combines variational autoencoders (VAEs) \cite{kingma2013auto} with CoGAN \cite{liu2016coupled}, a GAN framework where two generators share weights to learn the joint distribution of images in cross domains. CycleGAN \cite{zhu2017unpaired} and DiscoGAN \cite{kim2017learning} preserve key attributes between the input and the translated image by utilizing a cycle consistency loss. However, all these frameworks are only capable of learning the relations between two different domains at a time. Their approaches have limited scalability in handling multiple domains since different models should be trained for each pair of domains. Unlike the aforementioned approaches, our framework can learn the relations among multiple domains using only a single model.


\section{Star Generative Adversarial Networks} \label{stargan}

We first describe our proposed StarGAN, a framework to address multi-domain image-to-image translation within a single dataset. Then, we discuss how StarGAN  incorporates multiple datasets containing different label sets to flexibly perform image translations using any of these labels. 
\subsection{Multi-Domain Image-to-Image Translation} 
Our goal is to train a single generator $G$ that learns mappings among multiple domains. To achieve this, we train $G$ to translate an input image $x$  into an output image $y$ conditioned on the target domain label $c$, $G(x, c) \rightarrow y$. 
We randomly generate the target domain label $c$ so that $G$ learns to flexibly translate the input image. We also introduce an auxiliary classifier \cite{odena2016conditional} that allows a single discriminator to control multiple domains. That is,  our discriminator produces probability distributions over both sources and domain labels, $D: x \rightarrow \{{D}_{src}(x), {D}_{cls}(x)\}$. Fig.\thinspace\ref{model} illustrates the training process of our proposed approach.


\medskip

\noindent \textbf{Adversarial Loss.} To make the generated images indistinguishable from real images, we adopt an adversarial loss
\begin{equation}
\begin{split}
 \mathcal{L}_{adv} = & \thinspace {\mathbb{E}}_{x} \left[ \log{{D}_{src}(x)} \right]  \> \>  +   \\
 & \thinspace {\mathbb{E}}_{x, c}[\log{(1 - {D}_{src}(G(x, c)))}],
\end{split}
\label{eq1}
\end{equation}
\noindent where $G$ generates an image $G(x, c)$ conditioned on both the input image $x$ and the target domain label $c$, while $D$ tries to distinguish between real and fake images. In this paper, we refer to the term ${D}_{src}(x)$ as a probability distribution over sources given by $D$. The generator $G$ tries to minimize this objective, while the discriminator $D$ tries to maximize it.

\medskip

\noindent \textbf{Domain Classification Loss.} For a given input image $x$ and a target domain label $c$, our goal is to translate $x$ into an output image $y$, which is properly classified to the target domain $c$. To achieve this condition, we add an auxiliary classifier on top of $D$ and impose the domain classification loss when optimizing both $D$ and $G$. That is, we decompose the objective into two terms: a domain classification loss of real images used to optimize $D$, and a domain classification loss of fake images used to optimize $G$. In detail, the former is defined as
\begin{equation}
\mathcal{L}_{cls}^{r} = {\mathbb{E}}_{x, c'}[-\log{{D}_{cls}(c'|x)}],
\label{eq2}
\end{equation}
where the term ${D}_{cls}(c'|x)$ represents a probability distribution over domain labels computed by $D$. By minimizing this objective, $D$ learns to classify a real image $x$ to its corresponding original domain $c'$. We assume that the input image and domain label pair $(x, c')$ is given by the training data. On the other hand, the loss function for the domain classification of fake images is defined as
\begin{equation}
\mathcal{L}_{cls}^{f} ={\mathbb{E}}_{x, c}[-\log{{D}_{cls}(c|G(x, c))}].
\label{eq3}
\end{equation}
In other words, $G$ tries to minimize this objective to generate images that can be classified as the target domain $c$. 

\medskip

\noindent \textbf{Reconstruction Loss.} By minimizing the adversarial and classification losses, $G$ is trained to generate images that are realistic and classified to its correct target domain. However, minimizing the losses (Eqs.\thinspace\eqref{eq1} and \eqref{eq3}) does not guarantee that translated images preserve the content of its input images while changing only the domain-related part of the inputs. To alleviate this problem, we apply a cycle consistency loss \cite{kim2017learning, zhu2017unpaired} to the generator, defined as 
\begin{equation}
\mathcal{L}_{rec} = {\mathbb{E}}_{x, c, c'} [{||x - G(G(x, c), c')||}_{1} ],
\label{eq4}
\end{equation}
\noindent where $G$ takes in the translated image $G(x, c)$ and the original domain label $c'$ as input and tries to reconstruct the original image $x$. We adopt the L1 norm as our reconstruction loss. Note that we use a single generator twice, first to translate an original image into an image in the target domain and then to reconstruct the original image from the translated image. 

\medskip

\noindent \textbf{Full Objective.} Finally, the objective functions to optimize $G$ and $D$ are written, respectively, as \begin{equation}
\mathcal{L}_{D} =  - \mathcal {L}_{adv} +  {\lambda}_{cls}\thinspace\mathcal{L}_{cls}^{r}, 
\end{equation}
\begin{equation}
\mathcal{L}_{G} =   \mathcal {L}_{adv} +  {\lambda}_{cls}\thinspace\mathcal{L}_{cls}^{f} + 
{\lambda}_{rec}\thinspace\mathcal{L}_{rec},
\end{equation}
where ${\lambda}_{cls}$ and ${\lambda}_{rec}$ are hyper-parameters that control the relative importance of domain classification and reconstruction losses, respectively, compared to the adversarial loss. We use ${\lambda}_{cls} = 1$ and ${\lambda}_{rec} = 10$ in all of our experiments. 

\medskip

\subsection{Training with Multiple Datasets} \label{section 3.2}
An important advantage of StarGAN is that it simultaneously incorporates multiple datasets containing different types of labels, so that StarGAN can control all the labels at the test phase. An issue when learning from multiple datasets, however, is that the label information is only \textit{partially known} to each dataset. In the case of CelebA\thinspace\cite{liu2015faceattributes} and RaFD\thinspace\cite{langner2010presentation}, while the former contains labels for attributes such as hair color and gender, it does not have any labels for facial expressions such as `happy' and `angry', and vice versa for the latter. This is problematic because the complete information on the label vector $c'$ is required when reconstructing the input image $x$ from the translated image $G(x, c)$ (See Eq.\thinspace\eqref{eq4}). 

\medskip

\noindent\textbf{Mask Vector.} To alleviate this problem, we introduce a mask vector $m$ that allows StarGAN to ignore unspecified labels and focus on the explicitly known label provided by a particular dataset. In StarGAN, we use an $n$-dimensional one-hot vector to represent $m$, with $n$ being the number of datasets. In addition, we define a unified version of the label as a vector
\begin{equation}
\tilde{c} = [{c}_{1}, ..., {c}_{n}, m],
\label{eq7}
\end{equation}
\noindent where $[\cdot]$ refers to concatenation, and ${c}_{i}$ represents a vector for the labels of the $i$-th dataset. The vector of the known label ${c}_{i}$ can be represented as either a binary vector for binary attributes or a one-hot vector for categorical attributes. For the remaining $n \mathbb{-} 1$ unknown labels we simply assign zero values. In our experiments, we utilize the CelebA and RaFD datasets, where $n$ is two. 

\medskip

\begin{figure*}[ht]
\centering 
\centerline{\includegraphics[width=1.0\linewidth]{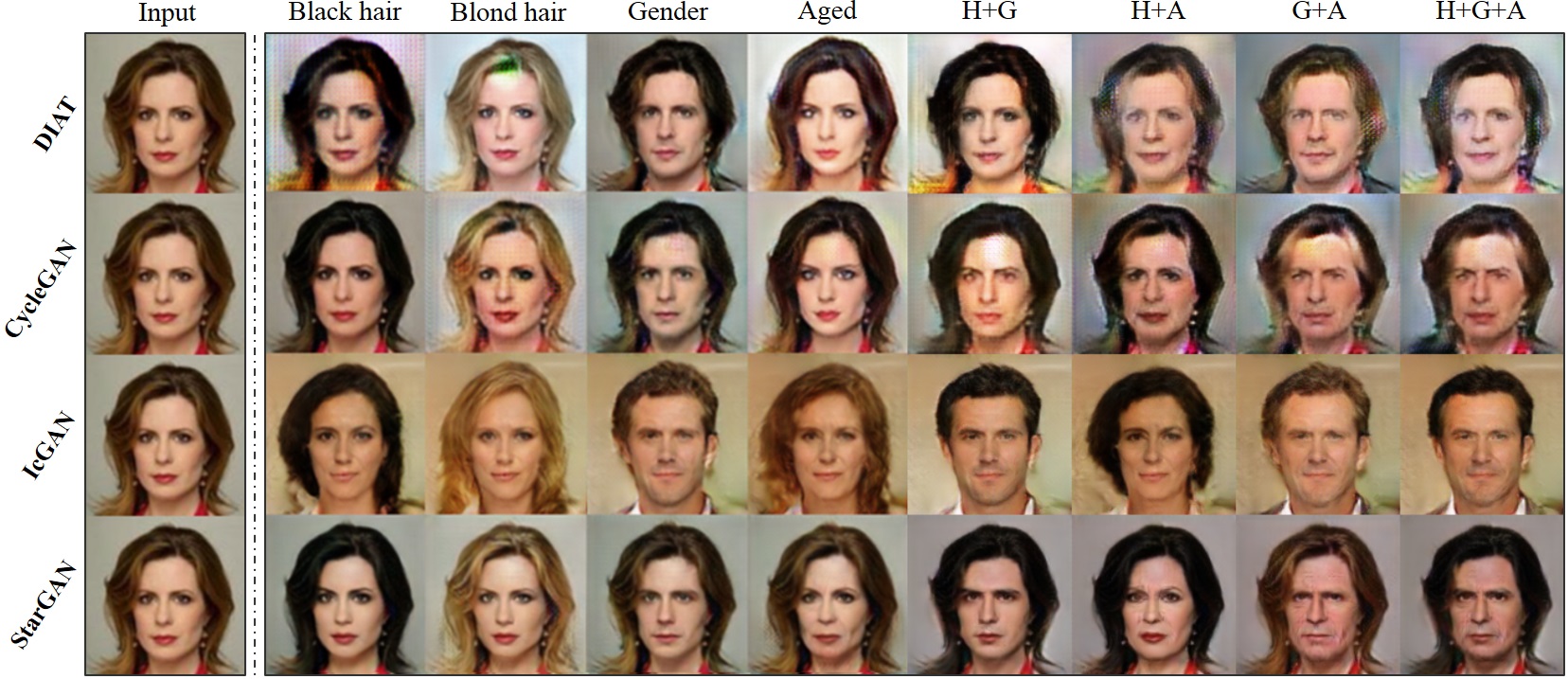}}
\caption{Facial attribute transfer results on the CelebA dataset. The first column shows the input image, next four columns show the single attribute transfer results, and rightmost columns show the multi-attribute transfer results. H: Hair color, G: Gender, A: Aged.}
\label{qual_celeb}

\end{figure*}

\noindent\textbf{Training Strategy.} When training StarGAN with multiple datasets, we use the domain label $\tilde{c}$ defined in Eq.\thinspace\eqref{eq7} as input to the generator. By doing so, the generator learns to \textit{ignore} the unspecified labels, which are zero vectors, and \textit{focus} on the explicitly given label. The structure of the generator is exactly the same as in training with a single dataset, except for the dimension of the input label $\tilde{c}$. On the other hand, we extend the auxiliary classifier of the discriminator to generate probability distributions over labels for all datasets. Then, we train the model in a multi-task learning setting, where the discriminator tries to minimize only the classification error associated to the known label. For example, when training with images in CelebA, the discriminator minimizes only classification errors for labels related to CelebA attributes, and not facial expressions related to RaFD. Under these settings, by alternating between CelebA and RaFD the discriminator learns all of the discriminative features for both datasets, and the generator learns to control all the labels in both datasets. 

\section{Implementation} \label{implementation}

\noindent \textbf{Improved GAN Training.}
To stabilize the training process and generate higher quality images, we replace Eq.\thinspace\eqref{eq1} with Wasserstein GAN objective with gradient penalty \cite{arjovsky2017wasserstein,gulrajani2017improved} defined as
\begin{equation}
\begin{split}
\mathcal{L}_{adv} = \thinspace & {\mathbb{E}}_{x}[{D}_{src}(x)] - {\mathbb{E}}_{x, c}[{D}_{src}(G(x, c))] \thinspace \thinspace\\  
& - {\lambda}_{gp} \thinspace {\mathbb{E}}_{\hat{x}}[{{(||{\triangledown}_{\hat{x}} {D}_{src}(\hat{x})||}_{2} - 1)}^{2}] \thinspace,
\end{split}
\end{equation}
where $\hat{x}$ is sampled uniformly along a straight line between a pair of a real and a generated images. We use ${\lambda}_{gp} = 10$ for all experiments.

\medskip

\noindent \textbf{Network Architecture.}
Adapted from CycleGAN \cite{zhu2017unpaired}, StarGAN has the generator network composed of two convolutional layers with the stride size of two for downsampling, six residual blocks \cite{he2016deep}, and two transposed convolutional layers with the stride size of two for upsampling. We use instance normalization \cite{ulyanov2016instance} for the generator but no normalization for the discriminator. We leverage PatchGANs \cite{Isola_2017_CVPR,li2016precomputed,zhu2017unpaired} for the discriminator network, which classifies whether local image patches are real or fake. See the appendix (Section \ref{section7_2}) for more details about the network architecture.
\begin{figure*}[ht]
\centering
\centerline{\includegraphics[width=1.0\linewidth]{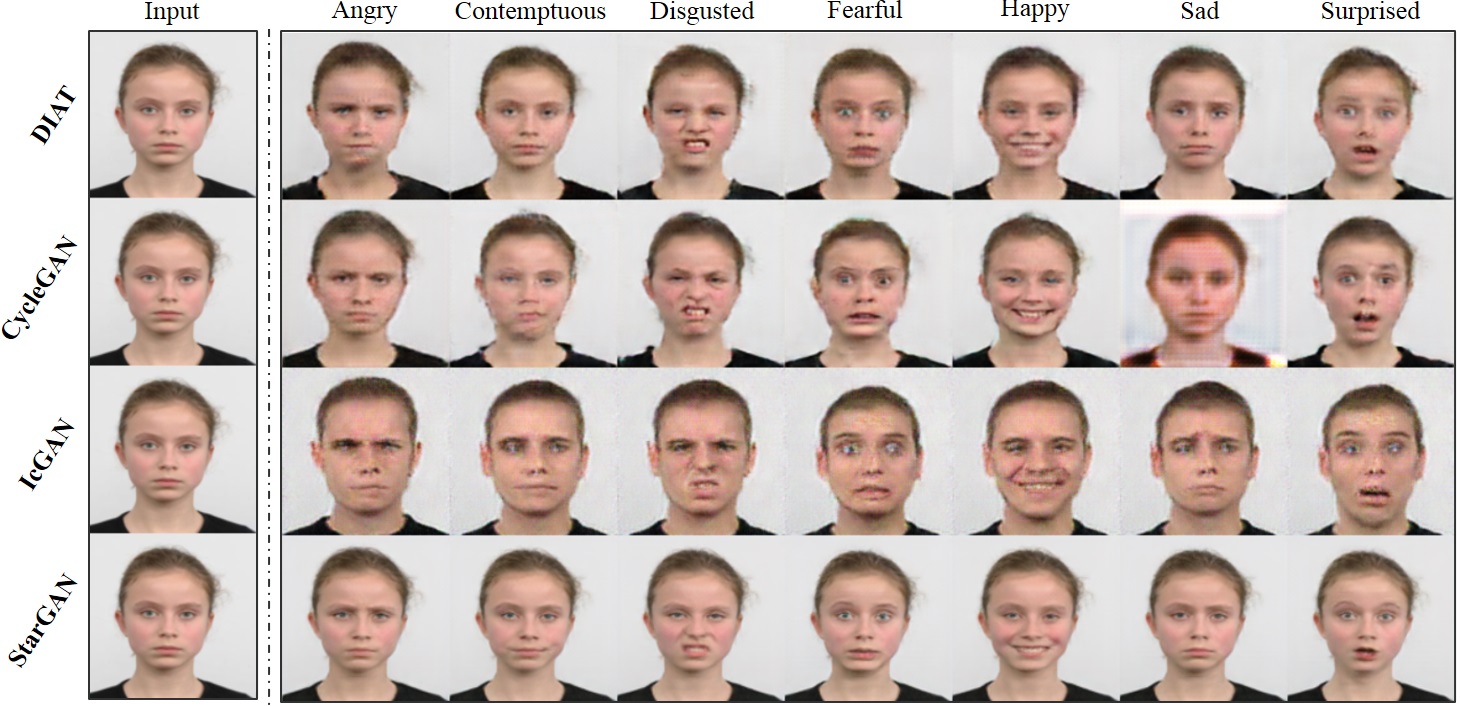}}
\caption{Facial expression synthesis results on the RaFD dataset.}
\label{qual_rafd}
\vspace{-0.1in}  
\end{figure*}

\section{Experiments} \label{experiments}

In this section, we first compare StarGAN against recent methods on facial attribute transfer by conducting user studies. Next, we perform a classification experiment on facial expression synthesis. Lastly, we demonstrate empirical results that StarGAN can learn image-to-image translation from multiple datasets. All our experiments were conducted by using the model output from unseen images during the training phase. 

\subsection{Baseline Models}
As our baseline models, we adopt DIAT \cite{li2016deep} and CycleGAN \cite{zhu2017unpaired}, both of which performs image-to-image translation between two different domains. For comparison, we trained these models multiple times for every pair of two different domains. We also adopt IcGAN \cite{perarnau2016invertible} as a baseline which can perform attribute transfer using a cGAN \cite{odena2016conditional}. 

\medskip

\noindent \textbf{DIAT} uses an adversarial loss to learn the mapping from $x \in X$ to $y \in Y$, where $x$ and $y$ are face images in two different domains $X$ and $Y$, respectively. This method has a regularization term on the mapping as ${||x - F(G(x))||}_{1} $ to preserve identity features of the source image, where $F$ is a feature extractor pretrained on a face recognition task. 

\medskip

\noindent \textbf{CycleGAN} also uses an adversarial loss to learn the mapping between two different domains $X$ and $Y$. This method regularizes the mapping via cycle consistency losses, ${||x - ({G}_{YX}({G}_{XY}(x)))||}_{1} $ and ${||y - ({G}_{XY}({G}_{YX}(y)))||}_{1}$. This method requires two generators and discriminators for each pair of two different domains.
\medskip

\noindent \textbf{IcGAN} combines an encoder with a cGAN \cite{odena2016conditional} model. cGAN learns the mapping $G: \{z, c\} \rightarrow x$ that generates an image $x$ conditioned on both the latent vector $z$ and the conditional vector $c$. In addition, IcGAN introduces an encoder to learn the inverse mappings of cGAN, ${E}_{z}: x \rightarrow z$ and ${E}_{c}: x \rightarrow c$. This allows IcGAN to synthesis images by only changing the conditional vector and preserving the latent vector.


\medskip

\subsection{Datasets}
\noindent \textbf{CelebA.} The CelebFaces Attributes (CelebA) dataset \cite{liu2015faceattributes} contains 202,599 face images of celebrities, each annotated with 40 binary attributes. We crop the initial $178 \times 218$ size images to $178 \times 178$, then resize them as $128 \times 128$. We randomly select 2,000 images as test set and use all remaining images for training data. We construct seven domains using the following attributes: hair color (\textit{black}, \textit{blond}, \textit{brown}), gender (\textit{male/female}), and age (\textit{young/old}). 

\medskip

\noindent \textbf{RaFD.} The Radboud Faces Database (RaFD) \cite{langner2010presentation} consists of 4,824 images collected from 67 participants. Each participant makes eight facial expressions in three different gaze directions, which are captured from three different angles. We crop the images to $256 \times 256$,  where the faces are centered, and then resize them to $128 \times 128$. 

\medskip

\subsection{Training}
All models are trained using Adam \cite{kingma2014adam} with ${\beta}_{1}=0.5$ and ${\beta}_{2}=0.999$. For data augmentation we flip the images horizontally with a probability of 0.5. We perform one generator update after five discriminator updates as in \cite{gulrajani2017improved}. The batch size is set to 16 for all experiments. For experiments on CelebA,  we train all models with a learning rate of 0.0001 for the first 10 epochs and linearly decay the learning rate to 0 over the next 10 epochs. To compensate for the lack of data, when training with RaFD we train all models for 100 epochs with a learning rate of 0.0001 and apply the same decaying strategy over the next 100 epochs. Training takes about one day on a single NVIDIA Tesla M40 GPU. 

\subsection{Experimental Results on CelebA}
We first compare our proposed method to the baseline models on a single and multi-attribute transfer tasks. We train the cross-domain models such as DIAT and CycleGAN multiple times considering all possible attribute value pairs. 
In the case of DIAT and CycleGAN, we perform multi-step translations to synthesize multiple attributes (e.g. transferring a gender attribute after changing a hair color).
\medskip

\begin{figure*}[t]
\centering
\centerline{\includegraphics[width=1.0\linewidth]{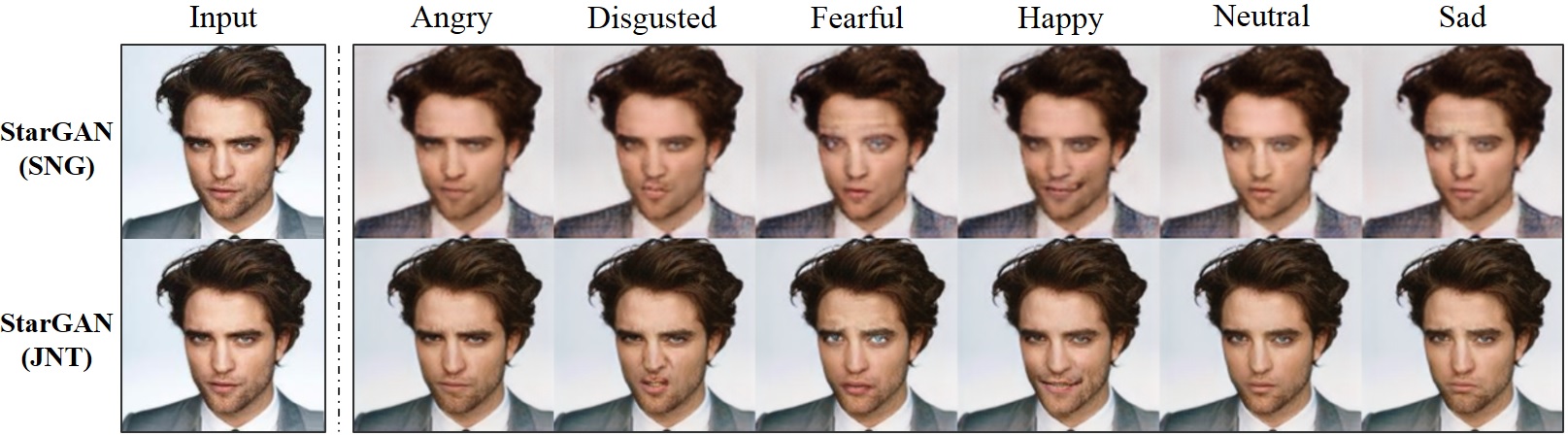}}
\caption{Facial expression synthesis results of StarGAN-SNG and StarGAN-JNT on CelebA dataset.}
\label{figure6}
\vspace{-0.1in}
\end{figure*}

\noindent\textbf{Qualitative evaluation.} Fig.\thinspace\ref{qual_celeb} shows the facial attribute transfer results on CelebA. We observed that our method provides a higher visual quality of translation results on test data compared to the cross-domain models. One possible reason is the regularization effect of StarGAN through a multi-task learning framework. In other words, rather than training a model to perform a fixed translation (e.g., brown-to-blond hair), which is prone to overfitting, we train our model to flexibly translate images according to the labels of the target domain. This allows our model to learn reliable features universally applicable to multiple domains of images with different facial attribute values. 

Furthermore, compared to IcGAN, our model demonstrates an advantage in preserving the facial identity feature of an input. We conjecture that this is because our method maintains the spatial information by using activation maps from the convolutional layer as latent representation, rather than just a low-dimensional latent vector as in IcGAN. 

\medskip

\noindent\textbf{Quantitative evaluation protocol.} For quantitative evaluations, we performed two user studies in a survey format using Amazon Mechanical Turk (AMT) to assess single and multiple attribute transfer tasks. Given an input image, the Turkers were instructed to choose the best generated image based on perceptual realism, quality of transfer in attribute(s), and preservation of a figure's original identity. The options were four randomly shuffled images generated from four different methods. The generated images in one study have a single attribute transfer in either hair color (black, blond, brown), gender, or age. In another study, the generated images involve a combination of attribute transfers. Each Turker was asked 30 to 40 questions with a few simple yet logical questions for validating human effort. The number of validated Turkers in each user study is 146 and 100 in single and multiple transfer tasks, respectively.
\medskip

\begin{table}[ht]
\setlength{\tabcolsep}{10pt}
\begin{center}
\begin{tabular}{l c c c c c}
Method & Hair color & Gender & Aged \\
\hline
DIAT & 9.3\% & 31.4\% & 6.9\% \\
CycleGAN & 20.0\% & 16.6\% & 13.3\% \\
IcGAN & 4.5\% & 12.9\% & 9.2\% \\
StarGAN & \textbf{66.2\%} & \textbf{39.1\%} & \textbf{70.6\%} \\
\hline
\end{tabular}
\end{center}
\caption{AMT perceptual evaluation for ranking different models on a single attribute transfer task. Each column sums to 100\%.}
\label{table1}
\vspace{-0.1in}
\end{table}

\begin{table}[ht]
\begin{center}
\begin{tabular}{l c c c c c}
Method & H+G & H+A & G+A & H+G+A \\
\hline
DIAT & 20.4\% &15.6\% & 18.7\% & 15.6\% \\
CycleGAN & 14.0\% & 12.0\% & 11.2\% & 11.9\% \\
IcGAN & 18.2\% & 10.9\% & 20.3\% & 20.3\% \\
StarGAN & \textbf{47.4\%} & \textbf{61.5\%} & \textbf{49.8\%} & \textbf{52.2\%} \\
\hline
a\end{tabular}
\end{center}
\caption{AMT perceptual evaluation for ranking different models on a multi-attribute transfer task. H: Hair color; G: Gender; A: Aged.}
\label{table2}
\vspace{-0.1in}
\end{table}

\noindent\textbf{Quantitative results.} Tables\thinspace\ref{table1} and \ref{table2} show the results of our AMT experiment on single- and multi-attribute transfer tasks, respectively. StarGAN obtained the majority of votes for best transferring attributes in all cases. 
In the case of gender changes in Table\thinspace\ref{table1}, the voting difference between our model and other models was marginal, e.g., 39.1\% for StarGAN vs. 31.4\% for DIAT. However, in multi-attribute changes, e.g., the `G+A' case in Table\thinspace\ref{table2}, the performance difference becomes significant, e.g., 49.8\% for StarGAN vs. 20.3\% for IcGAN), clearly showing the advantages of StarGAN in more complicated, multi-attribute transfer tasks. This is because unlike the other methods, StarGAN can handle image translation involving multiple attribute changes by randomly generating a target domain label in the training phase. 

\subsection{Experimental Results on RaFD}
We next train our model on the RaFD dataset to learn the task of synthesizing facial expressions. To compare StarGAN and baseline models, we fix the input domain as the `neutral' expression, but the target domain varies among the seven remaining expressions.

\medskip

\noindent\textbf{Qualitative evaluation.} As seen in Fig.\thinspace\ref{qual_rafd}, StarGAN clearly generates the most natural-looking expressions while properly maintaining the personal identity and facial features of the input. While DIAT and CycleGAN mostly preserve the identity of the input, many of their results are shown blurry and do not maintain the degree of sharpness as seen in the input. IcGAN even fails to preserve the personal identity in the image by generating male images. 

We believe that the superiority of StarGAN in the image quality is due to its implicit data augmentation effect from a multi-task learning setting. RaFD images contain a relatively small size of samples, e.g., 500 images per domain. When trained on two domains, DIAT and CycleGAN can only use 1,000 training images at a time, but StarGAN can use 4,000 images in total from all the available domains for its training. This allows StarGAN to properly learn how to maintain the quality and sharpness of the generated output.

\medskip

\noindent \textbf{Quantitative evaluation.} For a quantitative evaluation, we compute the classification error of a facial expression on synthesized images. We trained a facial expression classifier on the RaFD dataset (90\%/10\% splitting for training and test sets) using a ResNet-18 architecture \cite{he2016deep}, resulting in a near-perfect accuracy of 99.55\%. We then trained each of image translation models using the same training set and performed image translation on the same, unseen test set. Finally, we classified the expression of these translated images using the above-mentioned classifier. As can be seen in Table \ref{table3}, our model achieves the lowest classification error, indicating that our model produces the most realistic facial expressions among all the methods compared.

\begin{table}[ht]
\begin{center}
\begin{tabular}{l c c c}
Method & Classification error & \# of parameters \\
\hline
DIAT & 4.10 & 52.6M $\times$ 7 \\
CycleGAN & 5.99 & 52.6M $\times$ 14\\
IcGAN & 8.07 & 67.8M $\times$ 1\\
StarGAN & \textbf{2.12} & \textbf{53.2M $\times$ 1} \\
\hline
Real images & 0.45 & - \\
\end{tabular}
\end{center}
\caption{Classification errors [\%] and the number of parameters on the RaFD dataset. 
}
\label{table3}
\end{table}

Another important advantage of our model is the scalability in terms of the number of parameters required. The last column in Table \ref{table3} shows that the number of parameters required to learn all translations by StarGAN is seven times smaller than that of DIAT and fourteen times smaller than that of CycleGAN. This is because StarGAN requires only a single generator and discriminator pair, regardless of the number of domains, while in the case of cross-domain models such as CycleGAN, a completely different model should be trained for each source-target domain pair. 

\subsection{Experimental Results on CelebA+RaFD}

Finally, we empirically demonstrate that our model can learn not only from multiple domains within a single dataset, but also from \textit{multiple datasets}. We train our model jointly on the CelebA and RaFD datasets using the mask vector (see Section \ref{section 3.2}). To distinguish between the model trained only on RaFD and the model trained on both CelebA and RaFD, we denote the former as \textit{StarGAN-SNG} (single) and the latter as \textit{StarGAN-JNT} (joint). 

\smallskip

\noindent\textbf{Effects of joint training.} Fig.\thinspace\ref{figure6} shows qualitative comparisons between StarGAN-SNG and StarGAN-JNT, where the task is to synthesize facial expressions of images in CelebA. StarGAN-JNT exhibits emotional expressions with high visual quality, while StarGAN-SNG generates reasonable but blurry images with gray backgrounds. This difference is due to the fact that StarGAN-JNT learns to translate CelebA images during training but not StarGAN-SNG. In other words, StarGAN-JNT can leverage both datasets to improve shared low-level tasks such facial keypoint detection and segmentation. By utilizing both CelebA and RaFD, StarGAN-JNT can improve these low-level tasks, which is beneficial to learning facial expression synthesis. 


\smallskip

\begin{figure}[t]
\centering
\centerline{\includegraphics[width=1.0\linewidth]{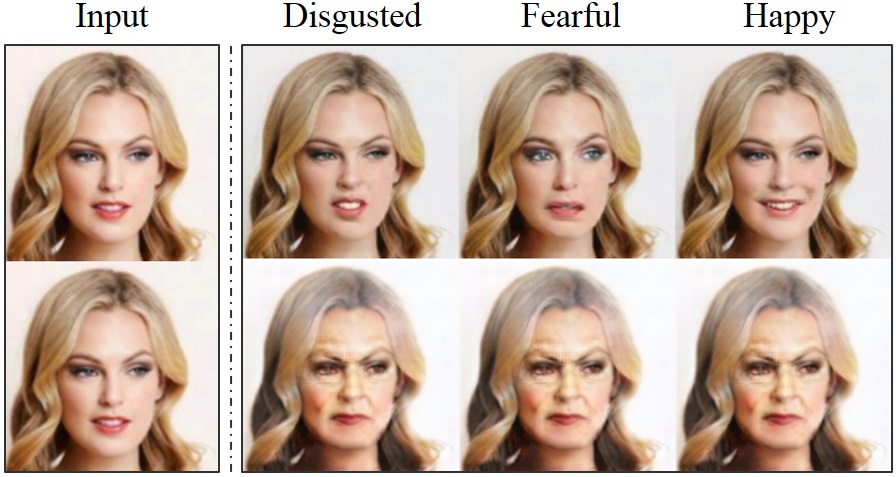}}
\caption{Learned role of the mask vector. All images are generated by StarGAN-JNT. The first row shows the result of applying the proper mask vector, and the last row shows the result of applying the wrong mask vector.}
\label{figure7}
\vspace{-0.1in}
\end{figure}

\noindent\textbf{Learned role of mask vector.} In this experiment, we gave a one-hot vector $c$ by setting the dimension of a particular facial expression (available from the second dataset, RaFD) to one. In this case, since the label associated with the second data set is explicitly given, the proper mask vector would be $[0,1]$. Fig.\thinspace\ref{figure7} shows the case where this proper mask vector was given and the opposite case where a wrong mask vector of $[1,0]$ was given. 
When the wrong mask vector was used, StarGAN-JNT fails to synthesize facial expressions, and it manipulates the age of the input image. This is because the model ignores the facial expression label as \textit{unknown} and treats the facial attribute label as \textit{valid} by the mask vector. Note that since one of the facial attributes is `young', the model translates the image from young to old when it takes in a zero vector as input. From this behavior, we can confirm that StarGAN properly learned the intended role of a mask vector in image-to-image translations when involving all the labels from multiple datasets altogether.

\section{Conclusion} \label{conclusion}

In this paper, we proposed StarGAN, a scalable image-to-image translation model among multiple domains using a single generator and a discriminator. Besides the advantages in scalability, StarGAN generated images of higher visual quality compared to existing methods~\cite{li2016deep,perarnau2016invertible,zhu2017unpaired}, owing to the generalization capability behind the multi-task learning setting. In addition, the use of the proposed simple mask vector enables StarGAN to utilize multiple datasets with different sets of domain labels, thus handling all available labels from them. We hope our work to enable users to develop interesting image translation applications across multiple domains. 

\vspace{7px}

\noindent\textbf{Acknowledgements.} This work was mainly done while the first author did a research internship at Clova AI Research, NAVER. We thank all the researchers at NAVER, especially Donghyun Kwak, for insightful discussions. This work was partially supported by the National Research Foundation of Korea
(NRF) grant funded by the Korean government (MSIP) (No. NRF2016R1C1B2015924). Jaegul Choo is the corresponding author.

{\small
\bibliographystyle{ieee}
\bibliography{egbib}
}
\vspace{10in}
\onecolumn
\section{Appendix} \label{appendix}
\subsection{Training with Multiple Datasets} 
\begin{figure*}[h]
\centering
\centerline{\includegraphics[width=0.95\linewidth]{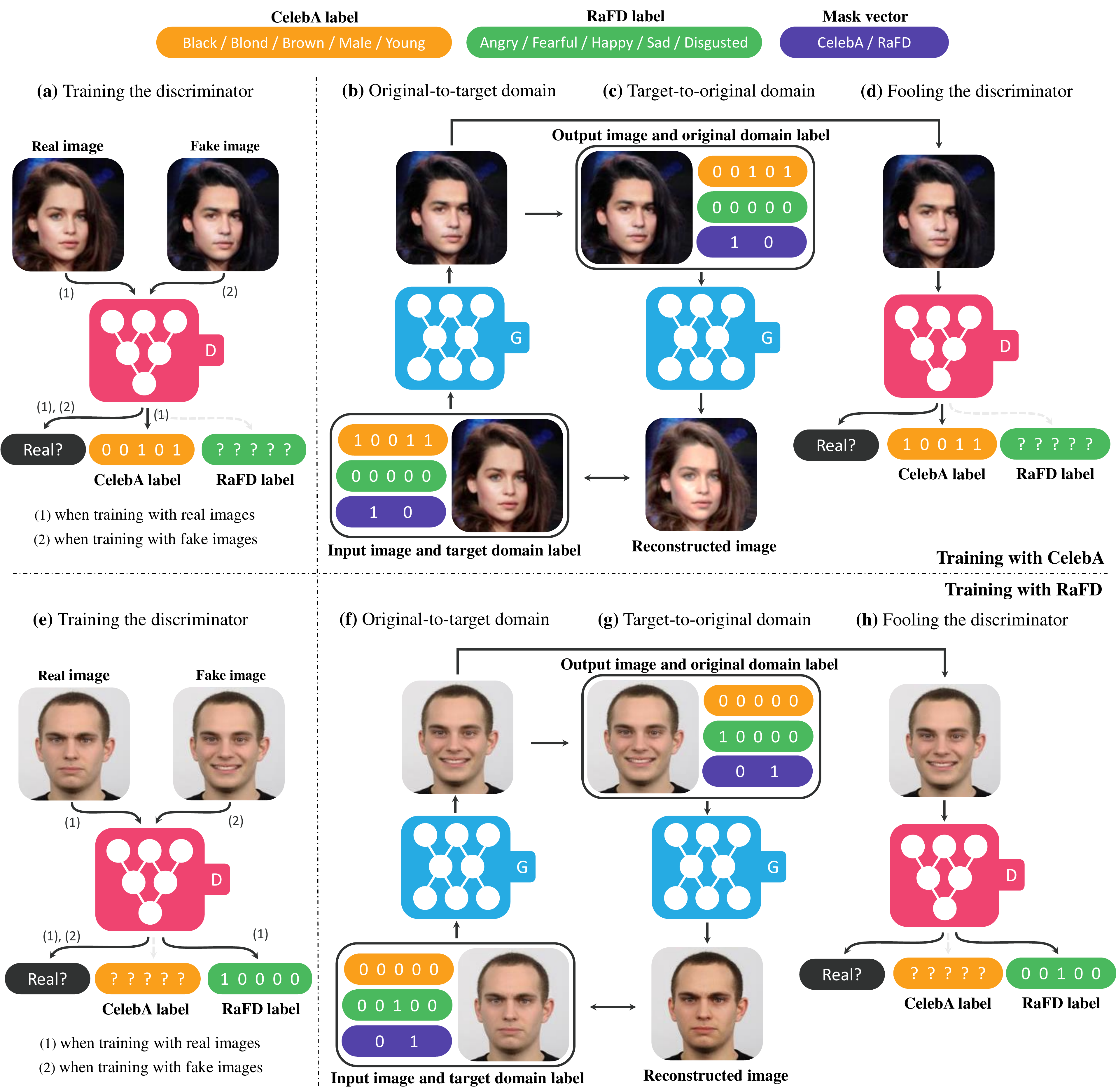}}
\medskip
\caption{Overview of StarGAN when training with both CelebA and RaFD. \textbf{(a) $\sim$ (d)} shows the training process using CelebA, and \textbf{(e) $\sim$ (h)} shows the training process using RaFD. \textbf{(a), (e)} The discriminator $D$ learns to distinguish between real and fake images and minimize the classification error only for the known label. \textbf{(b), (c), (f), (g)} When the mask vector (purple) is [1, 0], the generator $G$ learns to focus on the CelebA label (yellow) and ignore the RaFD label (green) to perform image-to-image translation, and vice versa when the mask vector is [0, 1]. \textbf{(d), (h)} $G$ tries to generate images that are both indistinguishable from real images and classifiable by $D$ as belonging to the target domain.}
\label{figure8}
\end{figure*}

Fig.\thinspace\ref{figure8} shows an overview of StarGAN when learning from both the CelebA and RaFD datasets. As can be seen at the top of the figure, the label for CelebA contains binary attributes (Black, Blond, Brown, Male, and Young), while the label for RaFD provides information on categorical attributes (Angry, Fearful, Happy, Sad, and Disgusted). The mask vector is a two-dimensional one-hot vector which indicates whether the CelebA or RaFD label is valid.

\medskip

\subsection{Network Architecture} \label{section7_2}
The network architectures of StarGAN are shown in Table \ref{table5} and \ref{table6}.  For the generator network, we use instance normalization in all layers except the last output layer. For the discriminator network, we use Leaky ReLU with a negative slope of 0.01. There are some notations; ${n}_{d}$: the number of domain, ${n}_{c}$: the dimension of domain labels (${n}_{d}+2$ when training with both the CelebA and RaFD datasets, otherwise same as ${n}_{d}$), N: the number of output channels, K: kernel size, S: stride size, P: padding size, IN: instance normalization.

\medskip

\begin{table*}[h]
\setlength{\tabcolsep}{13pt}
\renewcommand{\arraystretch}{1.7}
\begin{center}
\begin{tabular}{c  c  c}
Part & Input $\rightarrow$ Output Shape & Layer Information \\
\hline \hline
\multirow{3}{*}{Down-sampling} & $(h, w, 3+{n}_{c}) \rightarrow (h, w, 64)$ & CONV-(N64, K7x7, S1, P3), IN, ReLU \\
& $(h, w, 64) \rightarrow (\frac{h}{2}, \frac{w}{2}, 128)$ & CONV-(N128, K4x4, S2, P1), IN, ReLU \\
& $(\frac{h}{2}, \frac{w}{2}, 128) \rightarrow (\frac{h}{4}, \frac{w}{4},256)$ & CONV-(N256, K4x4, S2, P1), IN, ReLU \\
\Xhline{1.0pt}
\multirow{6}{*}{Bottleneck} & $(\frac{h}{4}, \frac{w}{4}, 256) \rightarrow (\frac{h}{4}, \frac{w}{4}, 256)$ & Residual Block: CONV-(N256, K3x3, S1, P1), IN, ReLU\\
 & $(\frac{h}{4}, \frac{w}{4}, 256) \rightarrow (\frac{h}{4}, \frac{w}{4}, 256)$ &  Residual Block: CONV-(N256, K3x3, S1, P1), IN, ReLU \\
 & $(\frac{h}{4}, \frac{w}{4}, 256) \rightarrow (\frac{h}{4}, \frac{w}{4}, 256)$ &   Residual Block: CONV-(N256, K3x3, S1, P1), IN, ReLU  \\
 & $(\frac{h}{4}, \frac{w}{4}, 256) \rightarrow (\frac{h}{4}, \frac{w}{4}, 256)$ &   Residual Block: CONV-(N256, K3x3, S1, P1), IN, ReLU  \\
 & $(\frac{h}{4}, \frac{w}{4}, 256) \rightarrow (\frac{h}{4}, \frac{w}{4}, 256)$ &   Residual Block: CONV-(N256, K3x3, S1, P1), IN, ReLU  \\
 & $(\frac{h}{4}, \frac{w}{4}, 256) \rightarrow (\frac{h}{4}, \frac{w}{4}, 256)$ &   Residual Block: CONV-(N256, K3x3, S1, P1), IN, ReLU  \\
\Xhline{1.0pt}
\multirow{3}{*}{Up-sampling} & $(\frac{h}{4}, \frac{w}{4}, 256) \rightarrow (\frac{h}{2}, \frac{w}{2}, 128)$ & DECONV-(N128, K4x4, S2, P1), IN, ReLU \\
 & $(\frac{h}{2}, \frac{w}{2}, 128) \rightarrow (h, w, 64)$ & DECONV-(N64, K4x4, S2, P1), IN, ReLU \\
 & $(h, w, 64) \rightarrow (h, w, 3)$ & CONV-(N3, K7x7, S1, P3), Tanh \\
\hline
\hline
\end{tabular}
\end{center}
\caption{Generator network architecture}
\label{table5}
\end{table*}

\medskip

\begin{table*}[h]
\setlength{\tabcolsep}{15pt}
\renewcommand{\arraystretch}{1.7}
\begin{center}
\begin{tabular}{c c c}
Layer & Input $\rightarrow$ Output Shape & Layer Information \\
\hline \hline
Input Layer & $(h, w, 3) \rightarrow (\frac{h}{2}, \frac{w}{2}, 64) $ & CONV-(N64, K4x4, S2, P1), Leaky ReLU \\
\Xhline{1.0pt}
Hidden Layer & $(\frac{h}{2}, \frac{w}{2}, 64) \rightarrow (\frac{h}{4}, \frac{w}{4}, 128) $ & CONV-(N128, K4x4, S2, P1), Leaky ReLU \\
Hidden Layer & $(\frac{h}{4}, \frac{w}{4}, 128) \rightarrow (\frac{h}{8}, \frac{w}{8}, 256) $ & CONV-(N256, K4x4, S2, P1), Leaky ReLU \\
Hidden Layer & $(\frac{h}{8}, \frac{w}{8}, 256) \rightarrow (\frac{h}{16}, \frac{w}{16}, 512) $ & CONV-(N512, K4x4, S2, P1), Leaky ReLU \\
Hidden Layer & $(\frac{h}{16}, \frac{w}{16}, 512) \rightarrow (\frac{h}{32}, \frac{w}{32}, 1024) $ & CONV-(N1024, K4x4, S2, P1), Leaky ReLU \\
Hidden Layer & $(\frac{h}{32}, \frac{w}{32}, 1024) \rightarrow (\frac{h}{64}, \frac{w}{64}, 2048) $ & CONV-(N2048, K4x4, S2, P1), Leaky ReLU \\
\Xhline{1.0pt}
Output Layer (${D}_{src}$) & $(\frac{h}{64}, \frac{w}{64}, 2048) \rightarrow (\frac{h}{64}, \frac{w}{64}, 1) $ & CONV-(N1, K3x3, S1, P1) \\
Output Layer (${D}_{cls}$) & $(\frac{h}{64}, \frac{w}{64}, 2048) \rightarrow (1, 1, {n}_{d}) $ & CONV-(N(${n}_{d}$), K$\frac{h}{64}$x$\frac{w}{64}$, S1, P0) \\
\hline \hline
\end{tabular}
\end{center}
\caption{Discriminator network architecture}
\label{table6}
\end{table*}

\medskip
\subsection{Additional Qualitative Results} 
\begin{figure*}[h]
\centering
\centerline{\includegraphics[width=1.0\linewidth]{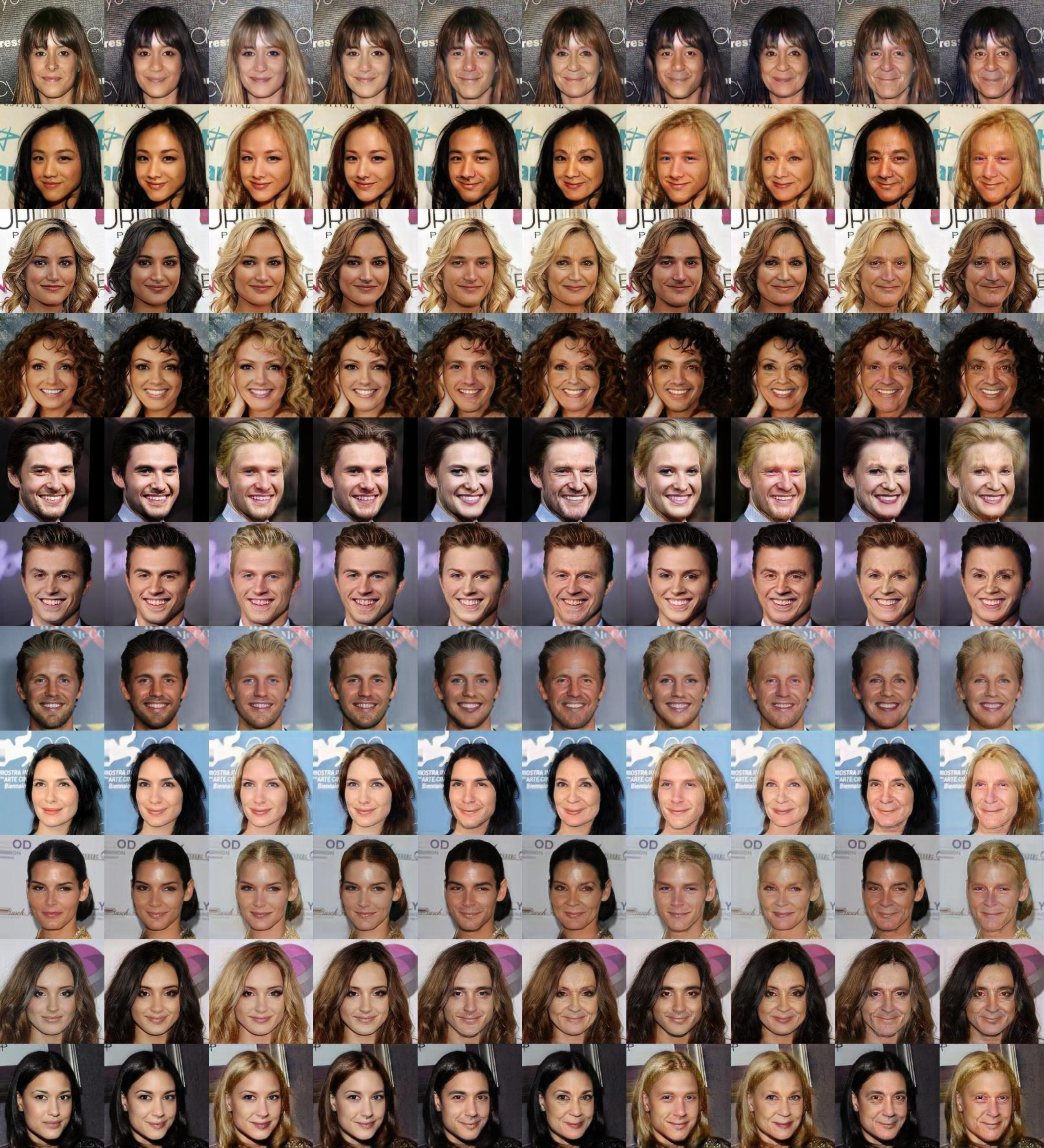}}
\caption{Single and multiple attribute transfer on CelebA (Input, Black hair, Blond hair, Brown hair, Gender, Aged, Hair color + Gender, Hair color + Aged, Gender + Aged, Hair color + Gender + Aged).}
\label{figure9}
\end{figure*}

\begin{figure*}[h]
\centering
\centerline{\includegraphics[width=1.0\linewidth]{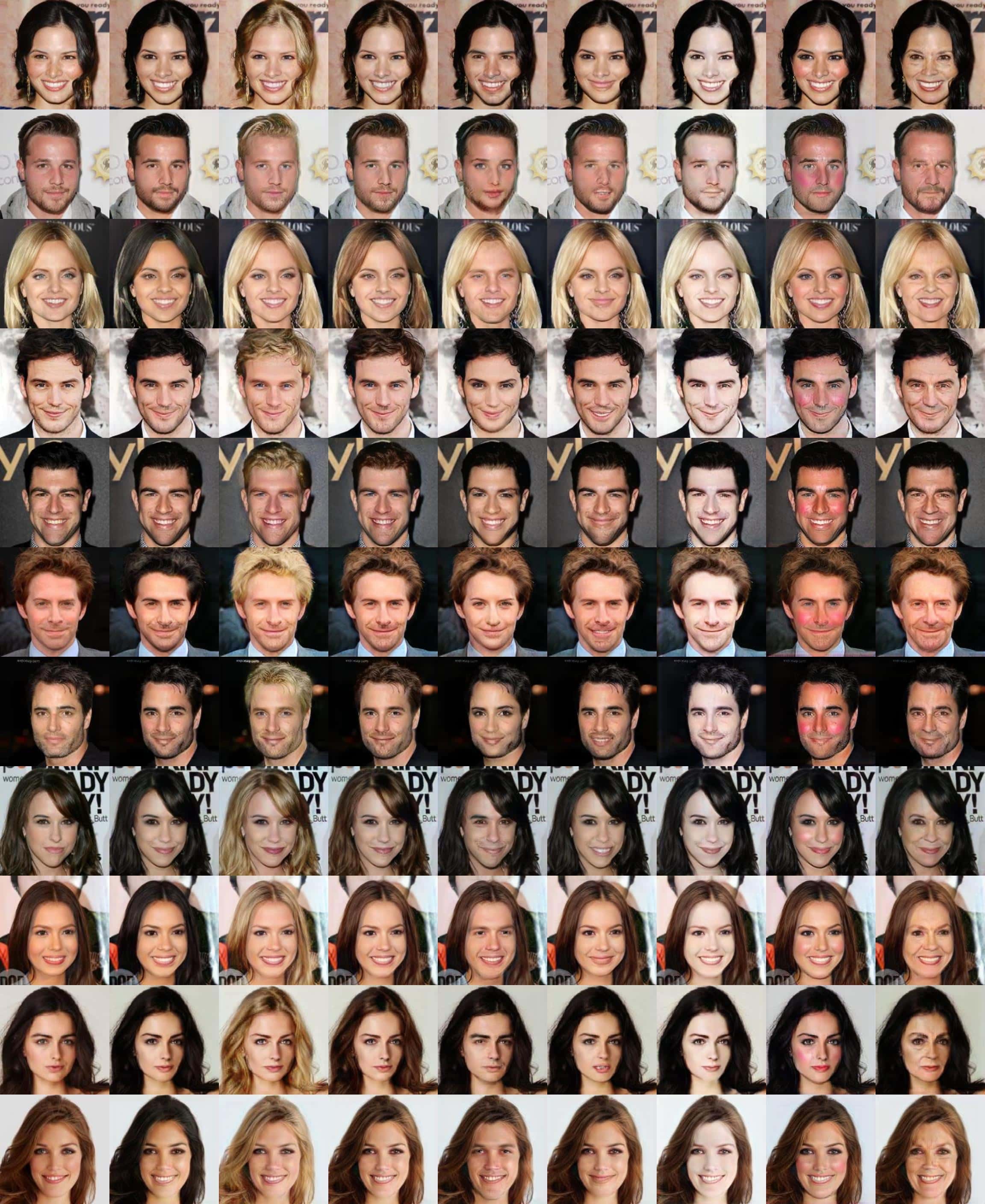}}
\caption{Single attribute transfer on CelebA (Input, Black hair, Blond hair, Brown hair, Gender, Mouth, Pale skin, Rose cheek, Aged).}
\label{figure10}
\end{figure*}

\begin{figure*}[h]
\centering
\centerline{\includegraphics[width=1.0\linewidth]{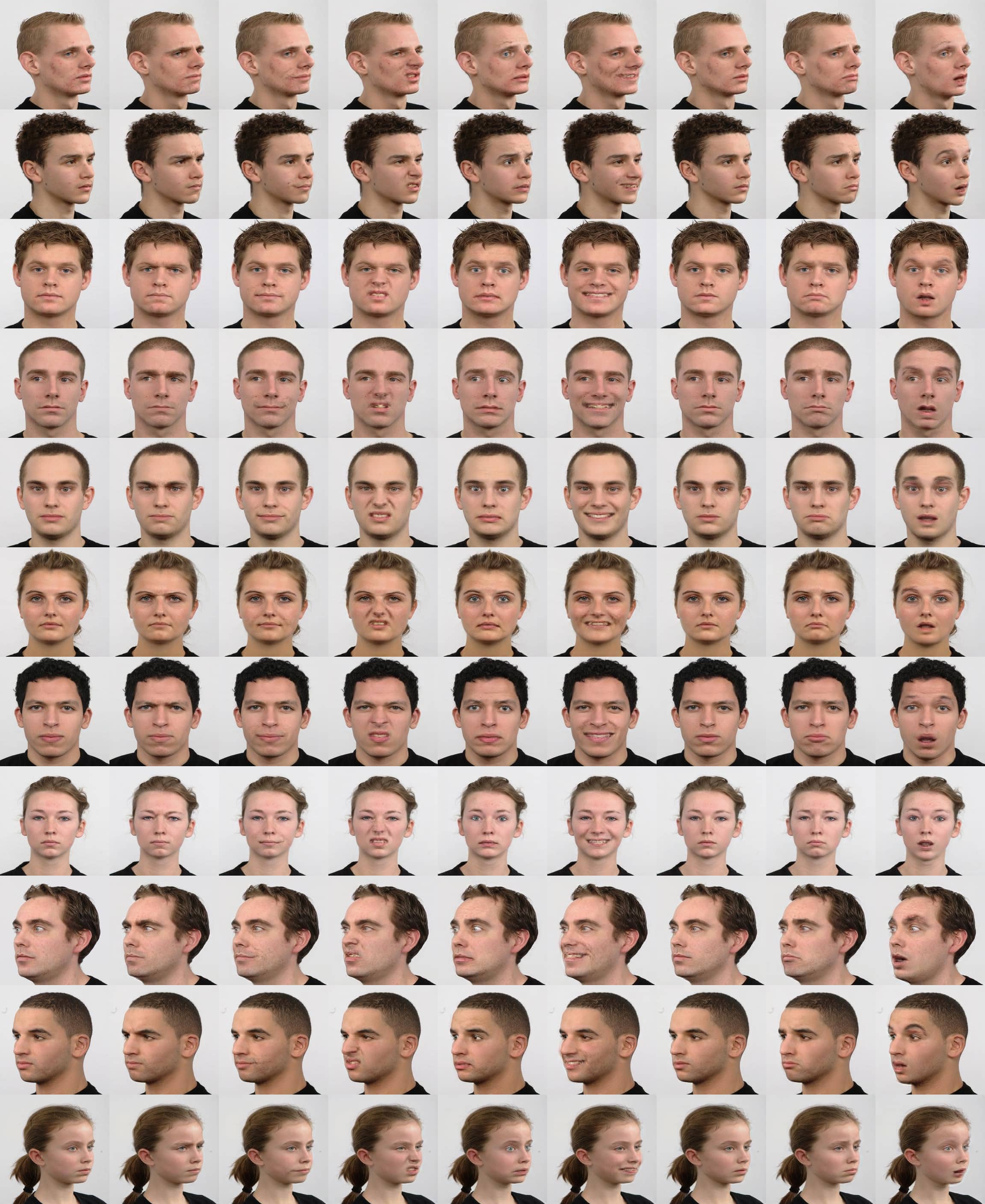}}
\caption{Emotional expression synthesis on RaFD (Input, Angry, Contemptuous, Disgusted, Fearful, Happy, Neutral, Sad, Surprised ).}
\label{figure11}
\end{figure*}

\begin{figure*}[h]
\centering
\centerline{\includegraphics[width=1.0\linewidth]{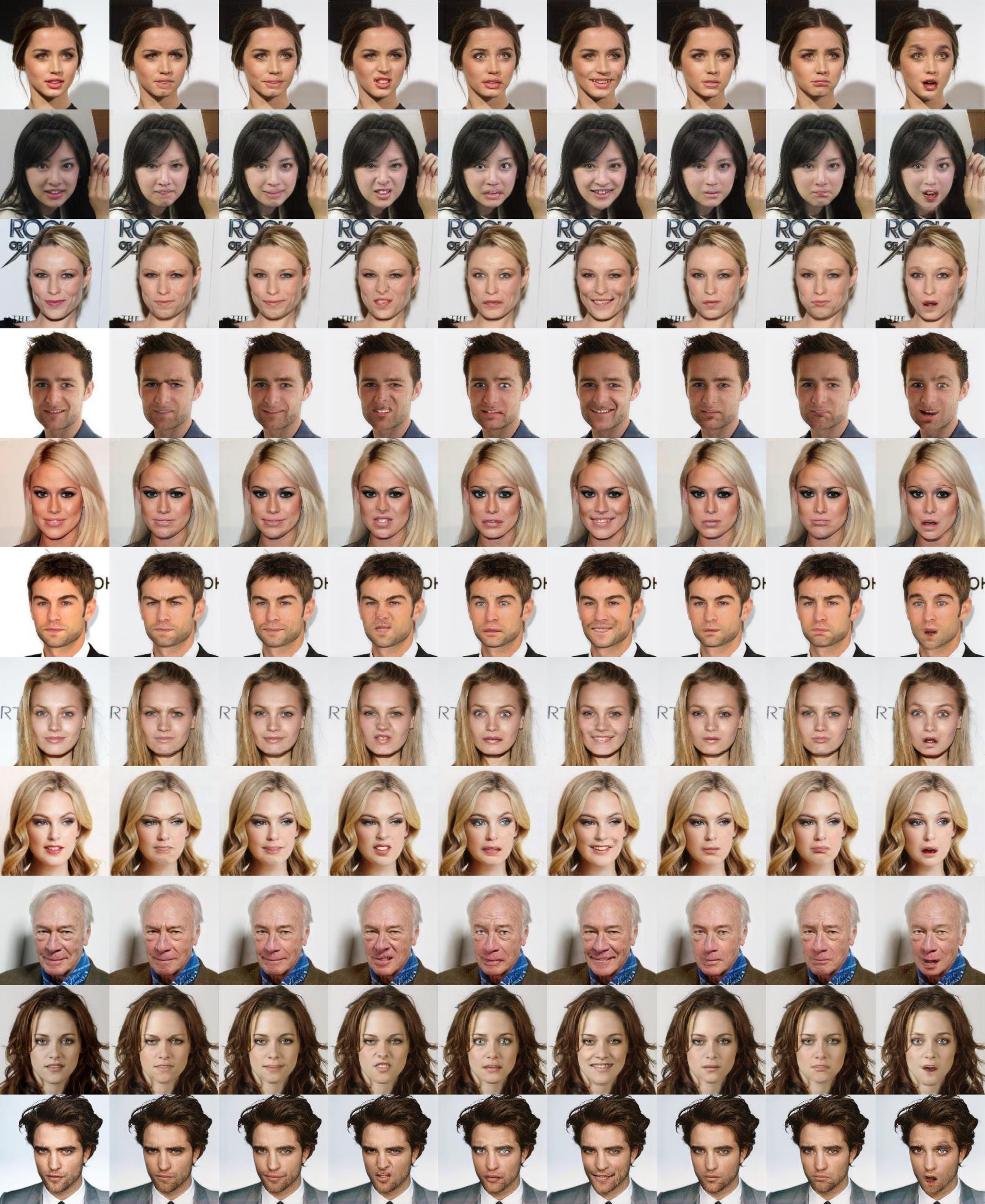}}
\caption{Emotional expression synthesis on CelebA (Input, Angry, Contemptuous, Disgusted, Fearful, Happy, Neutral, Sad, Surprised).}
\label{figure12}
\end{figure*}

Figs.\thinspace\ref{figure9},\thinspace\ref{figure10},\thinspace\ref{figure11}, and\thinspace\ref{figure12} show additional images with $256 \times 256$ resolutions generated by StarGAN. All images were generated by a single generator trained on both the CelebA and RaFD datasets. We trained StarGAN on a single NVIDIA Pascal M40 GPU for seven days.

\end{document}